\documentclass{Interspeech}

\interspeechcameraready

\title{Fact-Controlled Diagnosis of Hallucinations in Medical Text Summarization}

\author[]{Suhas}{BN} 
\author[]{Han-Chin}{Shing} 
\author[]{Lei}{Xu} 
\author[]{Mitch}{Strong} 
\author[]{Jon}{Burnsky} 
\author[]{Jessica}{Ofor} 
\author[]{Jordan R.}{Mason} 
\author[]{Susan}{Chen} 
\author[]{Sundararajan}{Srinivasan} 
\author[]{Chaitanya}{Shivade} 
\author[]{Jack}{Moriarty} 
\author[]{Joseph Paul}{Cohen} 

\affiliation{}{Amazon}{USA}
\email{}

\keywords{Hallucination detection, clinical dialogue summarization}

\usepackage{comment}

\begin{document}

\maketitle

\begin{abstract}

Hallucinations in large language models (LLMs) during summarization of patient-clinician dialogues pose significant risks to patient care and clinical decision-making.
However, the phenomenon remains understudied in the clinical domain, with uncertainty surrounding the applicability of general-domain hallucination detectors. 
The rarity and randomness of hallucinations further complicate their investigation. 
In this paper, we conduct an evaluation of hallucination detection methods in the medical domain, and construct two datasets for the purpose: A fact-controlled Leave-N-out dataset -- generated by systematically removing facts from source dialogues to induce hallucinated content in summaries; and a natural hallucination dataset -- arising organically during LLM-based medical summarization. 
We show that general-domain detectors struggle to detect clinical hallucinations, and that performance on fact-controlled hallucinations does not reliably predict effectiveness on natural hallucinations.
We then develop fact-based approaches that count hallucinations, offering explainability not available with existing methods. 
Notably, our LLM-based detectors, which we developed using fact-controlled hallucinations, generalize well to detecting real-world clinical hallucinations.
This research contributes a suite of specialized metrics supported by expert-annotated datasets to advance faithful clinical summarization systems. 
\end{abstract}

\section{Introduction}

The ability to generate precise summaries of patient-clinician dialogues and medical documentation is fundamental to healthcare delivery, enabling rapid comprehension of patient histories and informed clinical decision-making while reducing clinician's burnout \cite{moy2021measurement, quiroz2019challenges, 10.1001/jamanetworkopen.2019.9609}. Language models have enabled the development of automated clinical text summarization systems. However, a major challenge with these systems is the problem of ``hallucination'' - when the summary includes information that is not present in or misinterpreted from the original text. Hallucinations in clinical summaries pose significant risks to patient care, potentially leading to misdiagnosis, inappropriate treatment decisions, erosion of trust between patients and providers, as well as between providers and their software vendors. As healthcare increasingly adopts automated summarization tools, the ability to detect and prevent such hallucinations becomes crucial for ensuring patient safety and maintaining clinical accuracy.


The development of effective hallucination detection methods in clinical summarization is hindered by the lack of suitable datasets. Existing datasets often lack the specificity required for clinical contexts, do not provide controllable hallucinations, and may not capture the granular errors that can occur in medical summaries. 
To address these limitations, we propose the creation of a fact-controlled synthetic dataset that allows for precise control over the type and extent of hallucinations. We complement this synthetic dataset with another dataset consisting of natural hallucinations to ensure that we can measure performance in both a closed controllable setting and in real-world scenarios.

In this work, we systematically explore how existing methods detect hallucinations in clinical text summarization. We utilize two novel datasets created with clinical experts: Leave-N-Out (LNO) and Natural Hallucination (NH), which have omitted facts in different ways to exercise the capabilities of hallucination detection methods.
This paper aims to address the following research questions:
\noindent\textbf{RQ1:} How can we evaluate hallucination detection for medical tasks?
\noindent\textbf{RQ2:} Does the performance of hallucination detection metrics on synthetic data transfer to real-world clinical summarization data?
\noindent\textbf{RQ3:} How best can fact based hallucination detection be implemented?

To investigate these questions, we evaluate many baseline metrics, including classical, entailment-based, and LLM-based approaches on datasets described in \S\ref{sec:datasets}. Additionally, we develop many fact based methods for clinical summarization in \S\ref{sec:experiments}. Our findings demonstrate that LLM-based methods, particularly those using fact alignment, significantly outperform traditional metrics in hallucination detection. These methods show strong generalization, maintaining high performance across both synthetic and real-world datasets. 
Notably, our proposed methods also excel in detecting high-severity hallucinations, crucial for ensuring patient safety in clinical applications.

\section{Related Work}
\label{sec:relatedwork}
Traditional lexical and semantic overlap metrics such as ROUGE \cite{lin2004rouge}, BLEU \cite{papineni2002bleu}, and BERT-Score \cite{zhang2019bertscore} offer straightforward implementation but struggle with semantic understanding. More sophisticated approaches include entailment-based metrics (e.g., FactCC \cite{kryscinski2019evaluating}, SummaC \cite{laban2022summac}, AlignScore \cite{zha2023alignscore}), question-answering based metrics (e.g., FEQA \cite{durmus2020feqa}, QuestEval \cite{scialom2021questeval}, Q$^2$ \cite{honovich2021q}), and information theory-based metrics (e.g., InfoLM \cite{colombo2022infolm}). Recent developments include ensemble approaches like FENICE \cite{scire2024fenice} and LLM-based metrics such as DocLens \cite{xie2024doclens}. While DocLens represents an innovative approach to medical text evaluation, its methodology, which relies on claim generation and citation matching, doesn't directly address the need for hallucination detection in clinical summaries. Furthermore, its partial dependence on reference summaries conflicts with our goal of developing reference-free evaluation methods.

\section{Hallucination Datasets}
\label{sec:datasets}

We focus our research on the ACI-Bench dataset \cite{yim2023aci}, which comprises clinician-patient conversation transcripts and their corresponding clinical summaries (SOAP Notes \cite{podder2023soap}). A SOAP note is a structured way for healthcare providers to document patient information, including Subjective (patient reports), Objective (observable data), Assessment (provider's evaluation), and Plan (treatment strategy). The transcripts are generally longer and use layman's terms, while the summaries are written using medical terminology. 

\subsection{Leave-N-Out Dataset}
\label{sec:Leave-N-Out}

\begin{figure}[t]
\centering
  \includegraphics[width=1\columnwidth]{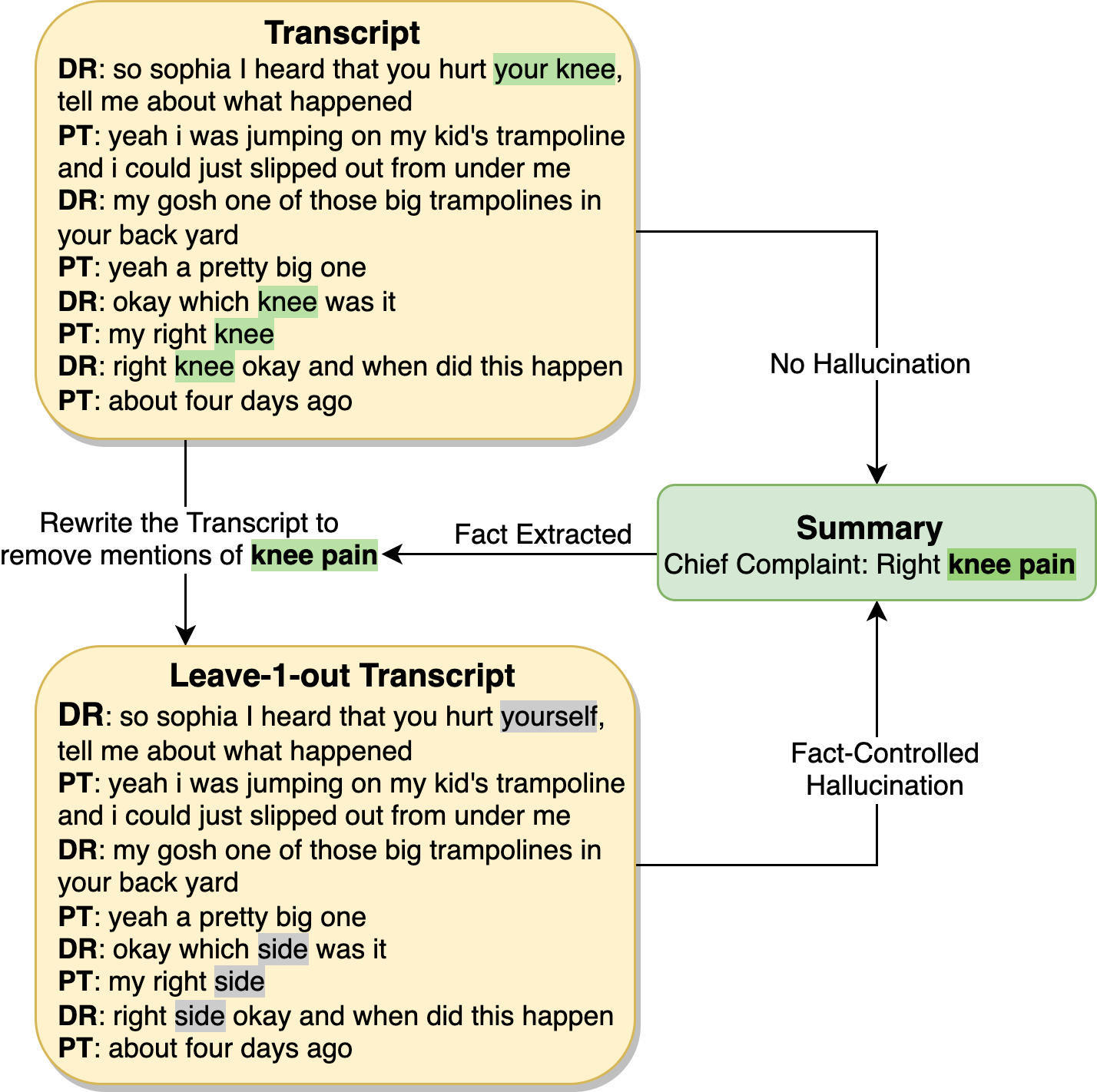}
  \caption{Dataset creation of the Leave-N-Out dataset, where a conversation transcript and its summary are shown side-by-side. The original transcript contains a dialogue about knee pain (in green). In the edited transcript, all references to knee pain are seamlessly replaced (in gray), while the summary remains unchanged and includes the specific detail `knee pain' (highlighted). This illustrates how the dataset systematically removes facts from transcripts while retaining them in summaries to simulate and study hallucinations in text summarization tasks, maintaining the meaning and natural flow of the conversation.}
  \label{fig:Leave-N-Out}
\end{figure}

In the Leave-N-Out dataset, we use an LLM (Sonnet 3.5 \cite{claude2024sonnet35}) to identify atomic facts from the summary and select N of them (such that they are orthogonal). We then use an LLM to rewrite the transcript to remove these occurrences while maintaining the meaning and natural flow of the conversation. For example, when N=0, it represents the original, unaltered ACI-Bench dataset, while N=1 indicates that all occurrences of a single fact have been removed from the transcript while leaving the summary unchanged. Refer to Figure \ref{fig:Leave-N-Out}. 
For example, complex medical statements are decomposed into atomic facts. The statement  ``Patient is a 35-year-old male with hypertension'' was broken down into three distinct facts: ``Patient is 35 years old'', ``Patient is male'', and ``Patient has hypertension''.
Removal of these facts would be considered N=3. Each fact is also annotated with a category of Age \& Sex, Exam Findings, Treatment Plan, Symptoms, Labs \& Imaging, Medical History, and Diagnosis and this is used to categorize high severity hallucinations as any category but Age \& Sex. 

These edited transcripts were then verified and manually edited by medical scribes to ensure each fact was correctly removed. In total 570 modified transcripts were generated, of these LLM modified transcripts only 428 lines needed to be corrected, meaning 0.75 lines per transcript required human correction.

\subsection{Natural Hallucination Dataset}
\label{sec:NaturalHallucination}
The Natural Hallucination (NH) dataset is a collection of model outputs that have hallucinations annotated by clinical scribes. The summaries are produced using varying levels of prompt complexity (simple, medium, and complex) to induce hallucinations in models of varying sizes (Claude Sonnet 3.5 \cite{claude2024sonnet35}, Mistral Large \cite{mistral2024large}, Llama 3.1 70B and 8B \cite{dubey2024llama}). 
Based on the LLM-generated summaries, clinical experts annotated and categorized each statement into four categories: Hallucination, Inference, Misunderstanding, and No Factual Error. We then aggregate these errors per subject and get an aggregated score for the errors (Hallucination +  Inference + Misunderstanding) while not considering the No Factual Error. This aggregated score is used as the value `N' in the NH dataset.

In addition to grouping errors, we identify high-severity categories, such as Diagnosis, Exam Findings, Lab Testing and Imaging, Medical History, Symptoms, and Treatment Plan, excluding Age and Sex. By focusing solely on these High Severity cases and disregarding low-severity instances, we enable a more focused examination of critical errors in medical information that could significantly impact patient care. The results of the High Severity correlation is in Table \ref{tab:correlation}.\footnote{A subset of the dataset has been made available online: \href{https://github.com/amazon-science/acibench-hallucination-annotations}{https://github.com/amazon-science/acibench-hallucination-annotations}}

\subsection{Cross-Domain Validation using XSum}
\label{sec:xsum}
To validate our methods beyond the medical domain, we also evaluate on XSum Hallucination Annotations \cite{maynez2020faithfulness}, a news summarization dataset known for its high hallucination rates. Unlike our medical datasets where hallucinations can impact clinical decision-making, XSum represents a different challenge where hallucinations may affect public information accuracy. 

The dataset consists of faithfulness and factuality annotations for 500 news articles, each paired with 5 different LLM generated summaries, resulting in 2,500 document-summary pairs with three independent judgements each. Annotators categorized hallucinations as intrinsic (information unsupported by the source) or extrinsic (information contradicting the source). This binary categorization contrasts with our fine-grained medical error taxonomy of hallucinations, inferences, and misunderstandings.

\section{Experiments}
\label{sec:experiments}
\subsection{Baseline Methods}
Our goal is to identify metrics that reliably quantify hallucinations in clinical text summaries. 
A hallucination-sensitive metric should show a monotonic relationship (increasing or decreasing) with the number of removed facts (N) in the LNO dataset. We use Pearson correlation between the model outputs and the number of removed facts to measure this relationship's strength. The absolute value of the correlation is used to allow us to compare between methods. Table \ref{tab:correlation} shows the results with the mean correlation ($\pm$ S.D.) across three trials. 
Bootstrapping is used to compute the standard deviation.
We can observe that most existing metrics perform poorly in detecting hallucinations, especially on the NH dataset. 

FENICE shows the best correlation (0.45) on LNO, suggesting sensitivity to the significant textual alterations in this dataset. However, it fails on NH (0.10), indicating limitations with subtle, natural hallucinations. SummaC$_{ZS}$ shows slight correlation on LNO (0.20) but is weak overall and drops significantly on NH. AlignScore achieves modest correlation on LNO (0.15) due to its focus on sentence-level entailment but struggles with the more complex NH dataset.

These findings highlight the limitations of existing summarization metrics for hallucination detection, particularly on the real world NH dataset. This underscores the need for new metrics specifically designed to address this challenge, motivating our development of novel metrics that focus on semantic meaning, factuality, and contextual consistency, as presented in the following sections.

\subsection{Is A Single Prompt All You Need?}
\label{sec:zs}

The axiomatic approach to identifying hallucinations involves checking each atomic unit of the summary against the transcript. Ideally, an LLM could directly provide a hallucination count (Single Prompt Counting). Alternatively, we can prompt an LLM to list unsupported atomic facts, which are then counted (Single Prompt List). Minor discrepancies (e.g., missing last names) are ignored, focusing on clinically relevant elements.  We use Claude Sonnet 3.5 for these and subsequent methods. Table \ref{tab:correlation} shows consistent performance across all datasets, with particularly strong performance on the Natural Hallucination dataset.

This suggests LLMs have an inherent ability to detect inconsistencies, including examples of factual errors. However, performance isn't perfect, potentially because we're overloading the single prompt with multiple tasks: understanding context, identifying facts, comparing them, and judging their validity. 

\begin{figure}[t]
    \centering
    \includegraphics[width=1\columnwidth]{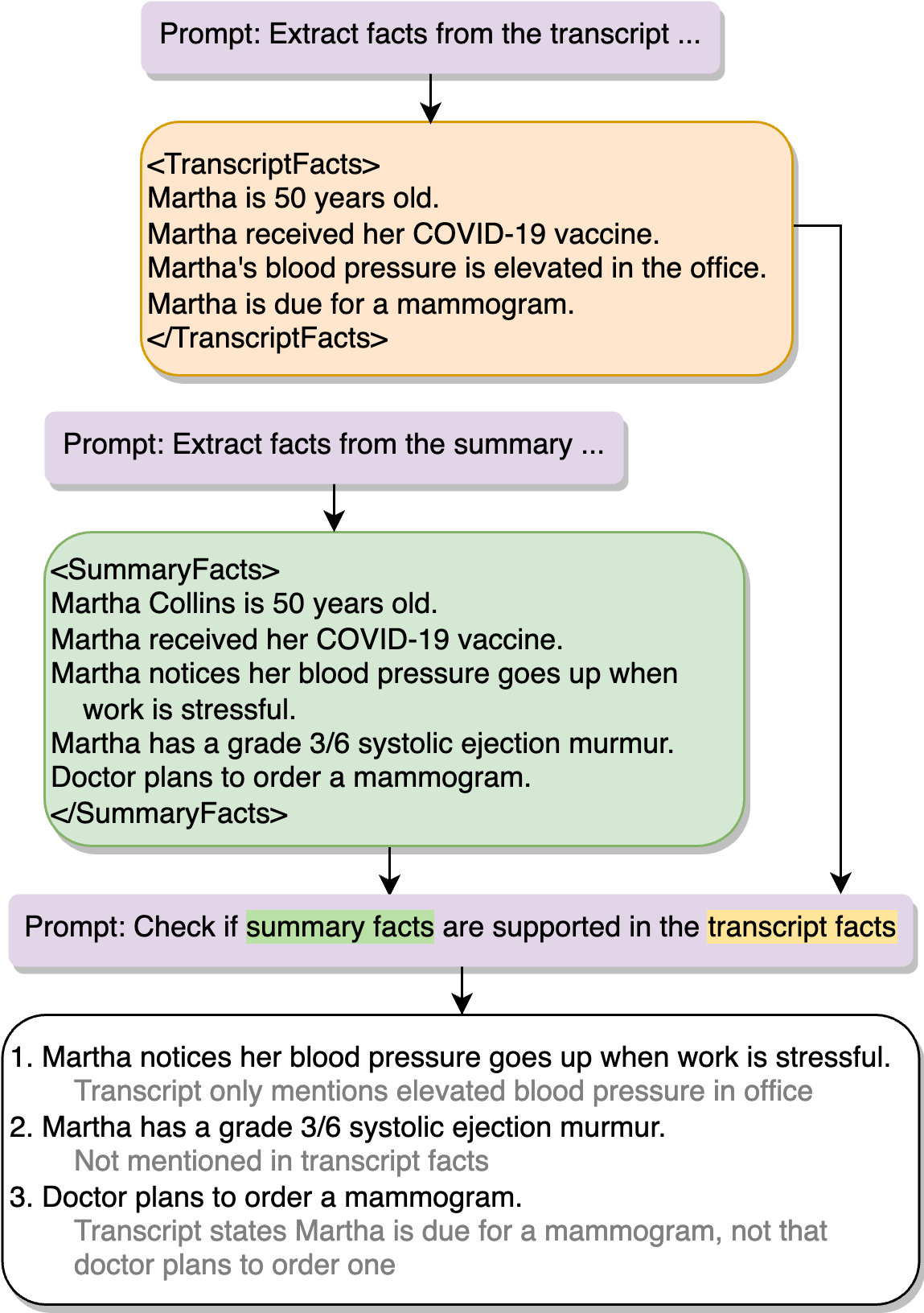}
    \caption{This figure illustrates the process of fact extraction and alignment using LLM prompting. Facts are first extracted separately from the transcript and summary. The LLM is then prompted to check if the summary facts are supported by the transcript facts, identifying discrepancies and potential unsupported information. This method allows for a detailed comparison of the extracted facts, highlighting areas where the summary may contain information not directly supported by the transcript.}
    \label{fig:fact-alignment-flow}
\end{figure}

\begin{table*}[t]
\caption{Comparison of Baseline, LLM-based, Semantic Similarity, and Fact-based Methods for Detecting Hallucinations in Clinical Text Summarization. Our methods demonstrate better generalization across datasets, maintaining high correlation scores even for clinically critical errors, as shown on the Leave-N-Out (LNO), Natural Hallucination (NH), and XSum  \cite{maynez2020faithfulness} datasets. High Severity columns highlight performance on errors deemed critical for clinical care.}
\label{tab:correlation}
\resizebox{\textwidth}{!}{%
\begin{tabular}{@{}lcc|cc|c@{}}
\toprule
\textbf{Metric} &
  \begin{tabular}[c]{@{}c@{}}Correlation \\LNO\end{tabular} &
  \multicolumn{1}{c}{\begin{tabular}[c]{@{}c@{}}Correlation \\LNO \\ High Severity\end{tabular}} &
  \begin{tabular}[c]{@{}c@{}}Correlation \\NH\end{tabular} &
  \multicolumn{1}{c}{\begin{tabular}[c]{@{}c@{}}Correlation\\ NH \\ High Severity\end{tabular}} &
  \multicolumn{1}{c}{\begin{tabular}[c]{@{}c@{}}Correlation\\ XSum \\ Hallucination\end{tabular}} \\ 
\midrule
ROUGE-1 \cite{lin2004rouge}                  & $0.08$$\pm$$0.04$                                                   & $0.13$$\pm$$0.04$ & $0.12$$\pm$$0.06$ & $0.03$$\pm$$0.05$ & $0.05$$\pm$$0.02$\\
FactCC \cite{kryscinski2019evaluating}                   &  $0.03$$\pm$$0.04$                                                    & $0.06$$\pm$$0.04$ & $0.01$$\pm$$0.04$ & $0.03$$\pm$$0.04$ & $0.15$$\pm$$0.02$\\
BERTScore \cite{zhang2019bertscore}                   & $0.01$$\pm$$0.04$                                               & $0.18$$\pm$$0.04$     & $0.16$$\pm$$0.06$ & $0.06$$\pm$$0.05$ &  $0.01$$\pm$$0.02$\\
QuestEval \cite{scialom2021questeval}                & $0.07$$\pm$$0.04$                                                    & $0.09$$\pm$$0.04$ & $0.07$$\pm$$0.06$ & $0.02$$\pm$$0.05$ & $0.00$$\pm$$0.02$\\
InfoLM$_{AB}$ \cite{colombo2022infolm}            & $0.02$$\pm$$0.04$                                                    & $0.08$$\pm$$0.04$ & $0.05$$\pm$$0.05$ & $0.06$$\pm$$0.04$ & $0.01$$\pm$$0.02$\\
SummaC$_{ZS}$ \cite{laban2022summac}               & $0.20$$\pm$$0.04$                                                   & $0.04$$\pm$$ 0.04$ & $0.05$$\pm$$0.04$ & $0.03$$\pm$$0.04$ & $0.10$$\pm$$0.02$\\
SummaC$_{Conv}$ \cite{laban2022summac}               & $0.03$$\pm$$0.04$                                                   & $0.14$$\pm$$0.04$ & $0.03$$\pm$$0.05$ & $0.04$$\pm$$0.06$ & $0.17$$\pm$$0.02$\\
AlignScore \cite{zha2023alignscore}               & $0.15$$\pm$$0.04$                                                    & $0.26$$\pm$$0.04$ & $0.06$$\pm$$0.05$ & $0.04$$\pm$$0.06$ & $0.28$$\pm$$0.02$\\
FENICE \cite{scire2024fenice}                   & \textbf{$0.45$$\pm$$0.03$}                                                    & \textbf{$0.43$$\pm$$0.03$} & $0.10$$\pm$$0.05$ & $0.01$$\pm$$0.04$ & $0.14$$\pm$$0.02$\\
\midrule
Single Prompt Counting            & $0.34$$\pm$$0.04$                                                    & $0.35$$\pm$$0.05$ & $0.19$$\pm$$0.06$ & $0.09$$\pm$$0.06$ & $0.15$$\pm$$0.02$\\
Single Prompt List            & $0.13$$\pm$$0.04$                                                    & $0.11$$\pm$$0.04$ & $0.23$$\pm$$0.07$ & $0.15$$\pm$$0.06$ & $0.16$$\pm$$0.02$\\

Sum. \& Transcript Fact Extract + Emb. Alignment (miniLM)         & $0.43$$\pm$$0.04$                                                    & $0.43$$\pm$$0.04$ & $0.32$$\pm$$0.10$ & $0.29$$\pm$$0.11$ & $0.13$$\pm$$0.02$ \\
Sum. \& Transcript Fact Extract + Emb. Alignment (bioBERT)          & $0.42$$\pm$$0.04$                                                    & $0.40$$\pm$$0.04$ & $0.37$$\pm$$0.00$ & $0.33$$\pm$$0.12$ & $0.16$$\pm$$0.02$ \\
Sum. \& Transcript Fact Extract + LLM  Alignment           & $0.43$$\pm$$0.04$                                                    & $0.41$$\pm$$0.05$ & $0.36$$\pm$$0.11$ & $0.34$$\pm$$0.15$ & $0.19$$\pm$$0.02$ \\
Sum. Fact Extract + Emb. Transcript Lookup (miniLM)            & $0.08$$\pm$$0.04$                                                    & $0.00$$\pm$$0.04$ & $0.19$$\pm$$0.08$ & $0.17$$\pm$$0.09$ & $0.24$$\pm$$0.02$  \\
Sum. Fact Extract + Emb. Transcript Lookup (bioBERT)            & $0.07$$\pm$$0.04$                                                    & $0.02$$\pm$$0.04$ & $0.20$$\pm$$0.09$ & $0.19$$\pm$$0.09$ & $0.24$$\pm$$0.02$ \\

Sum. Fact Extract + LLM Transcript Lookup           & $0.22$$\pm$$0.05$                                                    & $0.20$$\pm$$0.04$ & $0.16$$\pm$$0.10$ & $0.12$$\pm$$0.09$ & $0.19$$\pm$$0.02$ \\
\midrule
Semantic Emb. Similarity Score (miniLM)   & $0.30$$\pm$$0.04$                                                    & $0.35$$\pm$$0.04$ & $0.22$$\pm$$0.05$ & $0.13$$\pm$$0.06$ & $0.14$$\pm$$0.02$ \\
Semantic Emb. Similarity Score (bioBERT)  & $0.23$$\pm$$0.04$                                                    & $0.33$$\pm$$0.04$ & $0.15$$\pm$$0.06$ & $0.09$$\pm$$0.06$ &$0.17$$\pm$$0.02$\\

\bottomrule

\end{tabular}%
}
\end{table*}

\subsection{Is Chain-of-Prompts a better solution?}
\label{sec:zm}

We extend our methodology to a chain-of-prompts approach, breaking down the complex task into smaller steps. The first step remains the same: extracting concise, atomic, standalone facts from both the transcript and the summary. The second step involves aligning these facts to determine if facts from the summary are supported by facts from the transcript. We explain our approaches to this below:

\noindent\textbf{LLM Alignment} utilizes LLM prompting to identify and count unsupported facts from the lists of transcript and summary facts. Figure \ref{fig:fact-alignment-flow} details what this looks like. The lists of transcript and summary facts are included in the prompt and the output is list of unsupported facts which can be counted.

In Table \ref{tab:correlation} we observe improved performance over single prompt approaches.
Fact Alignment seems to work the best so far because it focuses on core facts, reducing noise (from other textual info) and simplifying analysis. By narrowing the scope and aligning key information, it minimizes errors and enhances efficiency. Its ability to handle partial matches captures  similarities, while the concise focus of facts makes inconsistencies easier to detect.

\noindent\textbf{Embedding Alignment} 
uses a sentence transformer 
to analyze transcript-summary fact pairs and compute their semantic similarity.  Embeddings for each fact from the summary and transcript are computed and all pairwise comparisons are computed, if a threshold is met then that fact from the summary is considered contained in the transcript. 

This comparison uses the cosine similarity and a threshold of 0.75 which was determined using a held-out dataset split. Embeddings are generated using pre-trained transformer models, all-MiniLM-L6-v2 \cite{wang2020minilm} (6 layers, 80MB) and BioBERT \cite{lee2020biobert} (12 layers, 400MB). While BioBERT offers domain-specific understanding, MiniLM provides comparable performance with faster inference and a smaller memory footprint. This allows us to investigate the trade-off between efficiency and domain expertise in capturing semantic similarity.

\noindent\textbf{Transcript Lookup} \label{sec:tl} uses the complete transcript instead of a list of extracted facts. The idea here is that fact extraction may remove context that is needed to match a fact. Potentially the fact extraction oversimplified a fact and now it appears to not match but given the context in the transcript it is not a hallucination.

As seen in Table \ref{tab:correlation}, using the full transcript leads to poorer results, contrary to our expectation. This suggests that context is important, but how it's incorporated into the evaluation process needs further investigation. It's possible that the current transcript lookup methods are not effectively leveraging the contextual information present in the full transcript.  We hypothesize that the fact alignment methods might be performing better because they reduce the search space and increase information density, allowing the LLM to focus on specific factual inconsistencies more efficiently. Further research is needed to explore alternative approaches for incorporating context that might improve the performance of transcript lookup methods.

\subsection{Semantic Similarity - Do we need to extract facts?}
\label{sec:semantic_sim}

As a ablation study we try to perform alignment without extracting facts, but by comparing each sentence from the summary with every sentence of the transcript. We first extract all sentences using spaCy \cite{honnibal2020spacy} and then compute embeddings and compute similarity as discussed in \S\ref{sec:zm}.

It is important to differentiate between the general BERTScore and our semantic similarity approach, which also uses sentence embeddings. While both methods leverage pre-trained models, our approach focuses on sentence-level comparisons, using the maximum similarity score between each summary sentence and all transcript sentences, along with a defined threshold (0.75). This stricter, sentence-level matching, combined with a precision-based metric, makes our method more sensitive to subtle hallucinations that might be missed by BERTScore's aggregate scoring.

\section{Conclusion}
\label{sec:conclusion_and_future_work}


We presented a novel fact-controlled approach to generate domain specific evaluation benchmarks using fact removal to rewrite the input in the LNO dataset. This approach makes it tractable to generate domain specific evaluation data when natural hallucinations are difficult to find in large numbers.
While this dataset was helpful in developing methods, high performance on this task didn't always translate into high performance on the real-world natural hallucination dataset.

Existing baseline methods show limited effectiveness on detecting clinical hallucinations, highlighting the need for specialized approaches.  
We also found that existing methods didn't generate an interpretable score as much as a count of hallucinations could and incorporated that into our approaches. We demonstrate fact based approaches offer clear explainability into what the hallucinations are as shown in Figure \ref{fig:fact-alignment-flow}.

Our proposed LLM-based evaluation methods, particularly fact alignment, generalize well across datasets and outperform existing metrics, achieving correlations up to 0.43 on synthetic data and 0.37 on real-world data. This suggests that decomposing evaluation into atomic fact verification provides a more robust framework for hallucination detection.

\section{Limitations}
The study primarily uses the ACI-Bench dataset, which may not represent the full complexity of clinical language. Additionally, our research focuses solely on SOAP notes, excluding other clinical documents like discharge summaries or progress reports. 

Fact based analysis is limited by the definition of what is a fact. While we tried to prompt models to extract atomic facts this often didn't work leading to undercounting of hallucinations.
This study also utilizes LLM-based evaluation methods which are computationally expensive, potentially limiting their practical application which may limit the utility.

\bibliographystyle{IEEEtran}
\bibliography{custom}

\end{document}